\documentclass[review]{elsarticle}

\usepackage{lineno,hyperref}
\modulolinenumbers[5]
\usepackage{bbm}
\usepackage{dsfont}

\usepackage[utf8]{inputenc} % allow utf-8 input
\usepackage[T1]{fontenc}    % use 8-bit T1 fonts
\usepackage{hyperref}       % hyperlinks
\usepackage{url}            % simple URL typesetting
\usepackage{booktabs}       % professional-quality tables
\usepackage{amsfonts}       % blackboard math symbols
\usepackage{nicefrac}       % compact symbols for 1/2, etc.
\usepackage{microtype}      % microtypography
\usepackage{lipsum}
\usepackage{amssymb}
\usepackage{xcolor}

\usepackage[a4paper, text={16cm,21cm},top=2.5cm, left=2.5cm, right=2.5cm, includeheadfoot, headheight=1cm]{geometry}

\usepackage{amsmath}

\usepackage{amsmath}

\usepackage{algorithm}
\usepackage[noend]{algpseudocode}
\usepackage{adjustbox}
\usepackage{booktabs} % To thicken table lines

\usepackage{amssymb}
\usepackage{multirow}
\usepackage{lscape}

\journal{Journal of \LaTeX\ Templates}

\begin{document}
\begin{frontmatter}
\title{Random Machines Regression Approach: an ensemble  support vector regression model with free kernel choice}

%% Group authors per affiliation:
\author{Anderson Ara}
\address{Department of Statistics\\
       Federal University of Bahia\\
       Salvador, Bahia, Brazil}
       
\author{Mateus Maia}
\address{Department of Statistics\\
       Federal University of Bahia\\
       Salvador, Bahia, Brazil}
       
\author{Samuel Mac\^edo}
\address{Department of Natural Sciences and Mathemathics\\
       Federal Institute of Pernambuco\\
       Recife, Pernambuco, Brazil}
                     
\author{ Francisco Louzada}
\address{Institute of Mathematical and Computer Sciences \\
       University of S\~ao Paulo \\
       S\~ao Carlos, S\~ao Paulo, Brazil}
%% or include affiliations in footnotes:
\begin{abstract}
Machine learning techniques always aim to reduce the generalized prediction error. In order to reduce it, ensemble methods present a good approach combining several models that results in a greater forecasting capacity. The Random Machines already have been demonstrated as strong technique, i.e: high predictive power, to classification tasks, in this article we propose an procedure to use the bagged-weighted support vector model to regression problems. Simulation studies were realized over artificial datasets, and over real data benchmarks. The results exhibited a good performance of Regression Random Machines through lower generalization error without needing to choose the best kernel function during tuning process.  \end{abstract}

\begin{keyword}
%% keywords here, in the form: keyword \sep keyword
\textit{Support Vector Regression \sep Bagging \sep Kernel Functions }%% MSC codes here, in the form: \MSC code \sep code
%% or \MSC[2008] code \sep code (2000 is the default)
\end{keyword}

\end{frontmatter}

%%
%% Start line numbering here if you want
%%
% \linenumbers

%% main text

\section{Introduction}
The prediction of new observations or events through statistical models is one of the main objectives of supervised statistical learning methods. Currently, machine learning models have several applications in regression tasks in a wide range of science fields, for instance, economy - predicting bitcoin's price \cite{mcnally2018predicting}, biology - predicting biological proprieties from plants \cite{feret2019estimating} or classifying gene functions \cite{park2008classification}, and physics - predicting electrical proprieties from materials. Inside this type of regression models, there is the support vector regression model that was proposed by \citep{drucker1997support} and has been used extensively as a optimal solution when compared with other traditional base-line methods \cite{xiao2018support, wu2004travel, delbari2019modeling,khosravi2018prediction}.  

Besides SVR, the ensemble method is a statistical learning approach that combines models in order to achieve greater predictive capacity. The combination of singular models can enhance predictive power and increase its generalization power \cite{van2007improved}. Even novel approaches, as deep learning models, can also benefit from ensemble procedures \cite{araque2017enhancing}. There are two main types of ensemble algorithms: bagging \cite{breiman1996bagging} that uses independent bootstrap samples to create multiple models and built a final classifier by average mean or majority vote, reducing the variance, and boosting algorithms \cite{freund1999short} that built sequential models in order to assign different weights based on their errors.

Bagging method does not require a specific type of base classifier, and can be be used to improve predictions in regression tasks \cite{rakesh2017ensemble,mendes2012ensemble,borra2002improving}. This method can be used to enhance the a single support vector regression model and others kind of algorithms. The bagging approach using support vector regression models is already reported in literature through diverse applications. As examples, can be cited the works: for predict protein retention time \cite{song2002prediction}, for predict time series \cite{deng2005ensemble}, electric load forecasting \cite{li2018subsampled}, for forecast building occupation \cite{wu2018using} and to predict blood pressure measures \cite{lee2018combining}. 

Nonetheless the different number of works that presents the bagging based on support vector regression models, there is no proposal of standard framework to choose which kernel function will be used in ensemble that use SVR as base-learners. In support vector models the kernel function and its hyperparameters have a compelling impact on the efficiency of the algorithm \cite{jebara2004multi}. Generally, this selection is made by a grid-search, which choose those parameters that produces the lower test error inside a grid of possible combinations, by random search \citep{bergstra2012random}, or by bayesian optimization algorithms \cite{bergstra2011algorithms}. All of them are computationally expensive and and can consume too much time. This work introduces a novel model that gives a solution for the kernel function's choice using the bagged supported vector regression with efficient computational time and robust predictive power named as the Regression Random Machines (RRM). The method received this name because it uses random kernel choice for each model that composes the bagged support vector regression method, besides propose weights to these regressors, increasing the predictive power of the final model. The result was validated over simulation studies, beyond using the algorithm on diverse benchmarking datasets.

The following chapter is organized on the ensuing outline. Section 3.2 presents a theoretical description about the support vector machine method, proposed by Vapnik \citep{drucker1997support}, the challenges on the selection of hyperparameters and standard kernel functions; Section 3.3 presents a overview of the bagging algorithm and bagged SVR; Section 3.4 introduce how the proposed Regression Random Machines (RRM) approach works in detail, followed by the simulations studies in Section 3.5, as well as the applications in real data in Section 3.6. In the last section, is proposed a discussion about correlation and strength of bagging base models, and how they affect the general method's performance .
\label{sec1}

\section{Support Vector Machine}
\begin{figure}[H]
    \centering
    \includegraphics{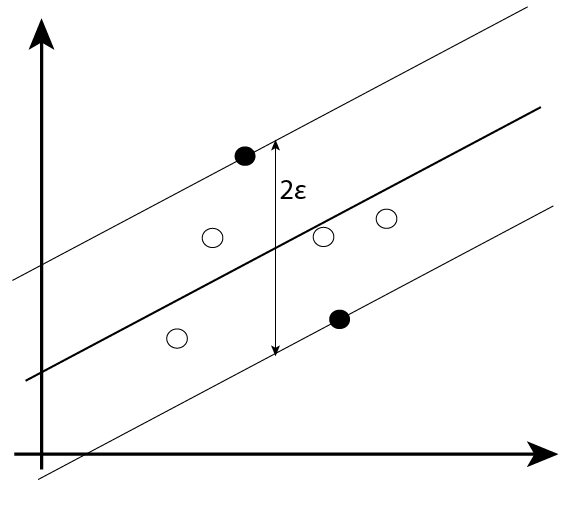}
    \caption{How support vector regression works: the goal its to build a maximum margin hypertube that maximizes the distance between the observations and the center of hypertube, which have a width of 2$\epsilon$}.
    \label{fig:svr_margin}
\end{figure}

Support vector regression machines \cite{drucker1997support} have been proposed as a generalized version of support vector machines \citep{cortes1995support} to regression tasks. Instead of creating a hyperplane with maximum margin in feature space, as done in support vector machines, the support vector regression build a optimal hyper-tube, which have all observations inside it, that maximizes the distance between the observations and the center of hypertube (Figure \ref{fig:svr_margin}). 

Supposing a database given by \{$\mathbf{x_{i}},\mathbf{y_{i}}$\}, i=1,$\dots$, $n$, $y_{i} \in \mathbb{R}$, where $n$ is the number of observations. The proposed objective function which will be minimized its given by 
\begin{equation}
    \min_{\mathbf{w}} \frac{1}{2} \mathbf{w} \boldsymbol{\cdot} \mathbf{w}
\end{equation}

\begin{equation*}
    \text{s.t}=  
\begin{cases}
  \mathbf{y}_{i}-\hat{f}(\mathbf{x}_{i}) \leq \epsilon \\
  \hat{f}(\mathbf{x}_{i}) - \mathbf{y}_{i} \leq \epsilon \\
\end{cases}
\end{equation*}

are satisfied for $i=1,\dots,n$; and where $\hat{f}(\mathbf{x}_{i})=\mathbf{w} \boldsymbol{\cdot} \mathbf{x}_{i} - b$. Vapnik \citep{drucker1997support} showed that there are cases where the hypertube cannot be created with all observations inside it, so they introduced the slacks variables $\mathcal{E}, \mathcal{E'}$ and the new optimization of convex objective function was given by

\begin{equation}
        \min_{\mathbf{w},\mathcal{E},\mathcal{E'}} \frac{1}{2} \mathbf{w} \boldsymbol{\cdot} \mathbf{w} + C\sum_{i=1}^{n}(\mathcal{E}+\mathcal{E'})
\end{equation}

\begin{equation*}
    \text{s.t}=  
\begin{cases}
  \mathbf{y}_{i}-\hat{f}(\mathbf{x}_{i}) \leq \mathcal{E} + \epsilon \\
  \hat{f}(\mathbf{x}_{i}) - \mathbf{y}_{i} \leq \mathcal{E}^\prime + \epsilon \\
  0 \geq \mathcal{E}, \mathcal{E}^\prime 
\end{cases}
\end{equation*}

for $i=1,\dots,n$ and hold $\hat{f}(\mathbf{x}_{i})=\mathbf{w} \boldsymbol{\cdot} \mathbf{x}_{i} - b$.

The hypertube also can be derived the dual to maximum-margin margin optimization given by the following Equation \ref{eq:dual_lagrange} and the constraints \cite{drucker1997support}

\begin{equation}
    \max_{\boldsymbol{\alpha,\alpha^{\prime}}} \left( \frac{1}{2} \sum_{i=1}^{n}\sum_{j=1}^{n} (\boldsymbol{\alpha}_{i}-\boldsymbol{\alpha^{\prime}}_{i})(\boldsymbol{\alpha}_{j}-\boldsymbol{\alpha^{\prime}}_{j})\mathbf{x}_{i} \boldsymbol{\cdot} \mathbf{x}_{j} + \sum_{i=1}^{n} (\boldsymbol{\alpha}_{i}-\boldsymbol{\alpha^{\prime}}_{i}) y_{i} - \epsilon \sum_{i=1}^{n}  (\boldsymbol{\alpha}_{i}-\boldsymbol{\alpha^{\prime}}_{i}) \right)
    \label{eq:dual_lagrange}
\end{equation}

\begin{equation*}
    \centering   
    \text{s.t}=  
    \begin{cases}
    \sum_{i=1}^{n}  (\boldsymbol{\alpha}_{i}-\boldsymbol{\alpha^{\prime}}_{i})=0 \\
      C \geq \boldsymbol{\alpha}_{i},\boldsymbol{\alpha^{\prime}}_{i} \geq 0 \\
    \end{cases}
\end{equation*}

for, $i=1,\dots,n$ where

    \begin{align*}
    & \boldsymbol{w}=(\boldsymbol{\alpha}_{i}-\boldsymbol{\alpha^{\prime}}_{i})\boldsymbol{y_{i}} \\
    & b=\frac{1}{n}\sum_{i=1}^{n}\sum_{j=1}^{n} 
    (\boldsymbol{\alpha}_{i}-\boldsymbol{\alpha^{\prime}}_{i})\mathbf{x}_{i} \boldsymbol{\cdot} \mathbf{x}_{j}-y_{i}
    \end{align*}

This approach of SVR works well with linear regression problems. However, there are cases where exist the non-linearity between the explaining and predictor variables. In these cases it may be used the kernel trick, based in Mercer's Theorem to deal with non-linearity. Using kernel methods, rather than consider the input space, it is considered higher feature spaces, where the observations could be linearly separable through the following function $K\mathbf{(x_{i},x_{j})}=\phi(\mathbf{x_{i}})\cdot\phi(\mathbf{x_{j}})$  that replaces the inner product in Equation \ref{eq:dual_lagrange}.

The functions $K(x,y)=\phi(\mathbf{x)}\cdot\phi(\mathbf{y})$ are defined as the semidefinite kernel functions \cite{courant1953methods}. Various types of kernel functions are used in distinct regression examples. The choice of particular kernels functions provide unique nonlinear mappings and the performance of the resulting SVR often depends on the appropriate choice of the kernel \cite{jebara2004multi}.  There are several kernel functions in the general framework for SVR, which some of the most commons were used in this paper. They are presented in Table \ref{tab:kernel_rrm}, and have hyperparameters $\gamma$ and $d$, which $\gamma \in \mathbb{R^{+}}, d \in \mathbb{N}$. The polynomial kernel, for instance, represent an transformation of the feature space to a determined degree. On the other hand, the radial basis kernel function as gaussian and laplacian, produces a feature space of infinite dimension.

\begin{table}[h]
\centering
\caption{Kernel Functions.}
\begin{tabular}{|l|c|c|}
\hline
\textbf{Kernel}   & \textbf{K(x,y)} & \textbf{Parameters}\\  \hline
Linear Kernel     & $\gamma(x\cdot y)$  & $\gamma$              \\
Polynomial Kernel & $(\gamma( x\cdot y))^{d}$ & $\gamma,d$   \\
Gaussian Kernel   & $e^{-\gamma||x-y||^2}$ & $\gamma$ \\
Laplacian Kernel  & $e^{-\gamma||x-y||}$ & $\gamma$ \\ \hline
\end{tabular}
\label{tab:kernel_rrm}
\end{table}
 
 \vspace{0.25cm}
 
 Determining which the best kernel by grid search, or other search method, can be an expensive and harrowing computational problem \citep{chapelle2000model}.  To solve it, many works have tried to develop a methodology which can improve the selection of the best kernel function \cite{jebara2004multi,ayat2005automatic,wu2009novel,friedrichs2005evolutionary,cherkassky2004practical}. Regression Random Machines method proposes an efficient alternative to work through a framework where it is avoidable this exhaustive search.

\section{Bagging}

Bagging is an acronym of Bootstrapping Aggregation, which was proposed by Breiman \cite{breiman1996bagging}. Bagging is an ensemble method that can be used for different prediction tasks. In general, the Bootstrapping Aggregating generates data sets by random sampling with replacement from the training set with the same size $n$, also known as bootstrap samples. Then, each model $h_{j}(x_{i})$ is trained independently for each bootstrapping sample $j$, $\forall j \in\{1,\dots,B\}$. The final bagging model, for regression tasks, is given by the following equation,
\begin{equation}
    H(\mathbf{x})=\frac{1}{n}\sum_{i=1}^{B}h_{i}(\mathbf{x}),
    \label{eq:bagging}
\end{equation}
where $h_{i}(\mathbf{x})$ is the model generated to each bootstrap sample from $i=1,,\dots,B$, and $B$ is the number of total bootstrap samples. 

Breiman \cite{breiman1996bagging} also reported that approximately $\frac{1}{3}$ of the observations from a database were not selected at each bootstrap sampling process. These observations were named as Out of Bag samples. Therefore, they could be used as test samples since they were not used to train the bootstrap models.

%it is  know that combining multiple classifiers, e.g ensemble methods, often increase the efficiency when compared to a single classifier \cite{duin1998combining}. 

\subsection{Bagging SVR}

Considering bagging procedure, the function $h_{i}(\mathbf{x})$ from (\ref{eq:bagging}) can be any model and it can improve the predictive power of non-parametric regression methods \cite{borra2002improving}. One possibility is to use the SVR as the base model in order to lower the generalization error. Besides the applications already shown, the use of the bagged SVR for regression tasks can be listed: content-based image retrieval \citep{yildizer2012efficient}, solar power forecasting \cite{abuella2017random}, quantifying urban land cover \citep{okujeni2016ensemble}, wind power prediction \citep{heinermann2014precise} and a trimmed bagging approach \citep{croux2007trimmed}. 

Despite the some works applied bagged SVR, none of them present a general framework to deal with the choice of the best kernel function. Often they choose it by trial evaluation, by a grid search or random search. As this proceeding is computationally expensive \cite{chapelle2000model}, this paper proposed a novel bagging approach that can overcome the difficult to choose the best kernel function, besides showing an improvement in the prediction capacity by combining several different SVR models and varying the kernel functions: the Regression Random Machines.  

\section{Regression Random Machines}
Given a training set $\{(x_{i},y_{i})\}_{i=1}^{N}$ with $\mathbf{x_{i}} \in \mathbb{R}^{p}$ and $y_{i} \in \mathbb{R}$, $\forall i=1,\dots,n$; the kernel bagging method initialize by training single models $h_{r}(\mathbf{x})$, where $r=1,\dots,R$, and $R$ is the  total number of different kernel functions that could be used in support vector regression models. For example, if $R=4$ a possible choice is define $h_{1}$ as SVR with \textit{Linear kernel} , $h_{2}$ as SVR with \textit{Polinomial kernel}, $h_{3}$ as SVR with \textit{Gaussian kernel} and $h_{4}$ as SVR with \textit{Laplacian kernel}. 

Each model is validated for the test set $\{(x_{k},y_{k})\}_{k=1}^{V}$, and the root mean square error ($RMSE_{r}$), which we will refer as $\delta_{i}$, is calculated for each model, $\forall r=1,\dots,R$, where $R$ is the number of kernel functions that will be used. As the range of the dependent variable in regression ($y$) is broad, the vector of root means squares $\boldsymbol{{\delta}}$ is divided by its deviation in order to standardize the error. Afterwards, sample probabilities, $\lambda_{r}$, are calculated by the Equation (\ref{eq:prob_rrm}) for each kernel function 

\begin{equation}
    \lambda_{r}=\frac{e^{-\beta \delta_{r}}}{\sum_{i=1}^{R}e^{-\beta \delta_{i}}}, 
    \label{eq:prob_rrm}
\end{equation}
with $\forall r=1,\dots,R$.

Subsequently, $B$ bootstrap samples are sampled from the training set. A support vector regression model $g_{k}$ is trained for each bootstrap sample, $k=i,\dots,B $ and the kernel function that will be used for $g_{k}$ will be determined by a random choice with probability $\lambda_{r}, \forall r=1,\dots,R$.
The probabilities $\lambda_{r}$ are higher if determined kernel function used in $h_{r}(\mathbf{x})$ has lower generalization error measured from the calculated RMSE over the test set. Consequently, the models with lower RMSE will frequently appear when the random kernel selection for each bootstrap model is done. 

\begin{figure}[H]
    \centering
    \includegraphics[width=0.6\textwidth]{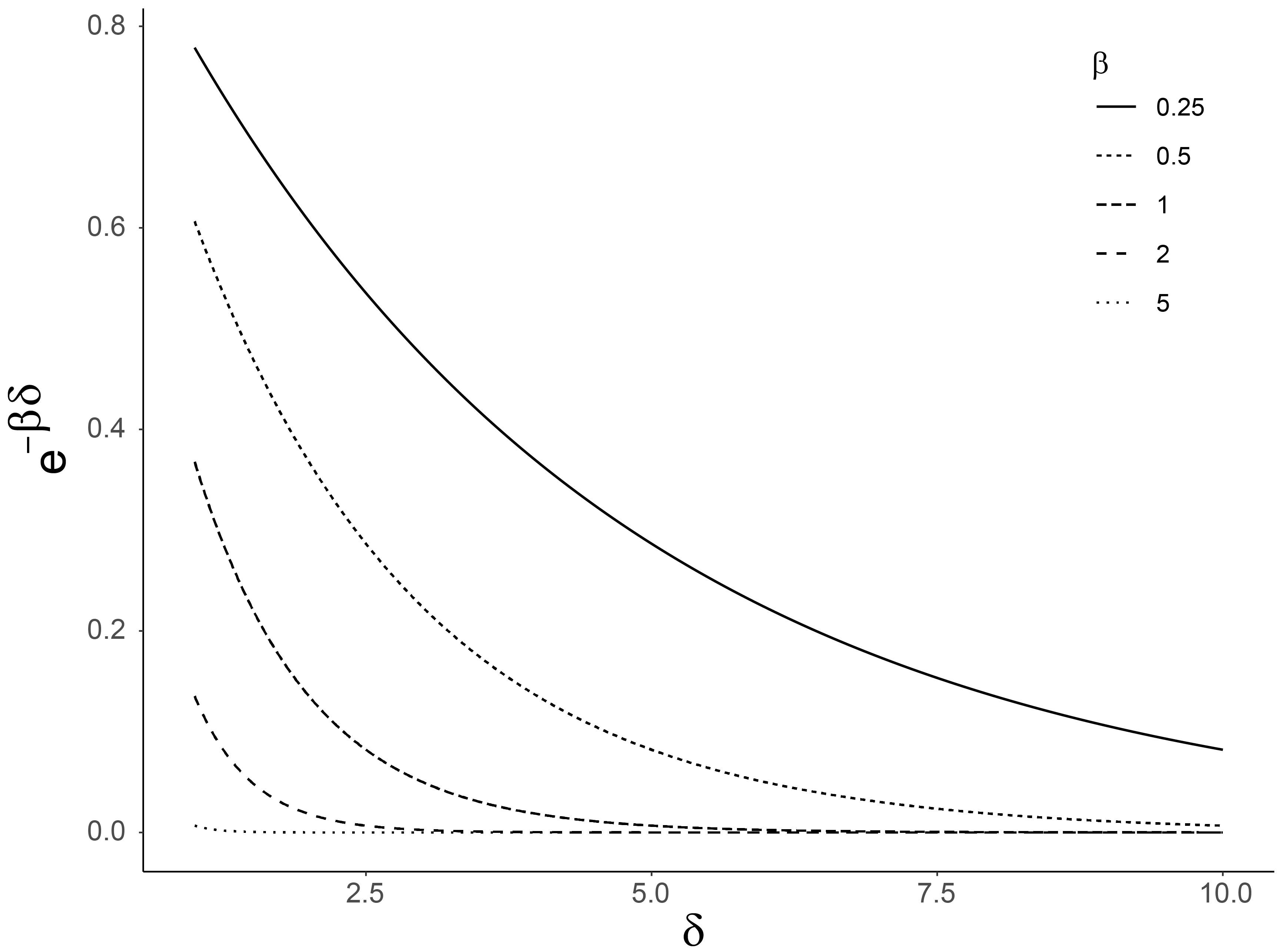}
    \caption{The correlation coefficient and its relation with the calculation of the probabilities $\boldsymbol{\lambda}$. As $\beta$ increase the penalty given by $\delta$ values decrease.}
    \label{fig:cof_corr}
\end{figure}

The parameter $\beta$, named as correlation parameter, will tune the penalty of the generalization error of each model. Figure \ref{fig:cof_corr} shows that small values of $\beta$ create heavy-tail penalty functions while greater beta's values represent light-tail penalty. The parameter gets its name because it can determine the diversity between the chosen kernel functions, since high values further penalize the performance differences between each SVR model types. For instance, considering a value of $\beta=0$, the result of vector of probabilities is given by Equation \ref{eq:prob_rrm} its $\boldsymbol{\lambda}=\{0.25,0.25,0.25,0.25\}$, which means that all kernels have the same chance to be sampled in each bootstrap model, i.e: maximum diversity between kernel functions.  On other hand, a large value for $\beta$ would quickly scale difference between models, consequently, the vector of probabilities $\boldsymbol{\lambda}$ would accumulate in a single kernel, and just one would be sampled, i.e: minimum diversity since its the same kernel function for all bootstrap samples returning to the traditional bagging approach.

After, a weight $w_{i}$ is assigned to each bootstrap model calculated for $g_{i}$ $\forall i=1,\dots,B$. The weight is given by the Equation (\ref{eq:weight_rrm}).

\begin{equation}
  w_{i}=\frac{e^{-\beta \Lambda_{i}}}{\sum_{j=1}^{B}e^{-\beta \Lambda_{j}}}, \hspace{0.3cm} i=1,\dots,B ,
    \label{eq:weight_rrm}
\end{equation}

where $\Lambda_{i}$ is the Root Mean Square Error of model's prediction $g_{i}$ using the Out of Bag Samples ($OOBG_{i}$), obtained from $i$ bootstrap sample $\forall i=1,\dots,B$, as test set.

%\begin{equation}
%    \Lambda_{i}=\frac{1}{T_{i}}\sum_{k=1}^{T_{i}}(g_{i}(\boldsymbol{x_{k}})-y_{k})^{2} ,
%    \label{eq:rmse_oob}
%\end{equation}

%where $T_{i}$ is the \# of observations of each $OOBG_{i}$.

\begin{equation}
    G(\mathbf{x_{i}})= \sum_{j}^{B}w_{j}g_{j}(\mathbf{x_{i}}), \hspace{0.3cm} i=1,\dots,N.
    \label{eq:final_class}
\end{equation}

All the modeling process is summed up in the pseudo-code exposed in Algorithm \ref{alg:RM}.

\begin{algorithm}[H]

\caption{Random Machines}
\begin{algorithmic} 
    \small
    \State{\bf{Input}: Training Data, Test Data, B, Kernel Functions}
    \For{each $Kernel Function_{r}$}
        \State Calculate the model  $h_{r}$
        %\State Calculate the Root Mean Squared Error $\delta_{r}$
    \EndFor
    \State \textbf{Calculate} the probabilities $\lambda_{r}$
    \State \textbf{Generate} B bootstrap samples
    \For{b in B}
        \State Model $g_{b}( \mathbf{x_{i}})$ by sampling a kernel function with probability
        $\lambda_{r}$
        \State Assign a weight $w_{b}$ using $OOBG_{b}$ samples.
    \EndFor
    \State \textbf{Calculate} $G(\mathbf{x})$
\end{algorithmic}
\label{alg:RM}

\end{algorithm}

\begin{figure}[H]
    \centering
    \includegraphics[width=0.45\textwidth]{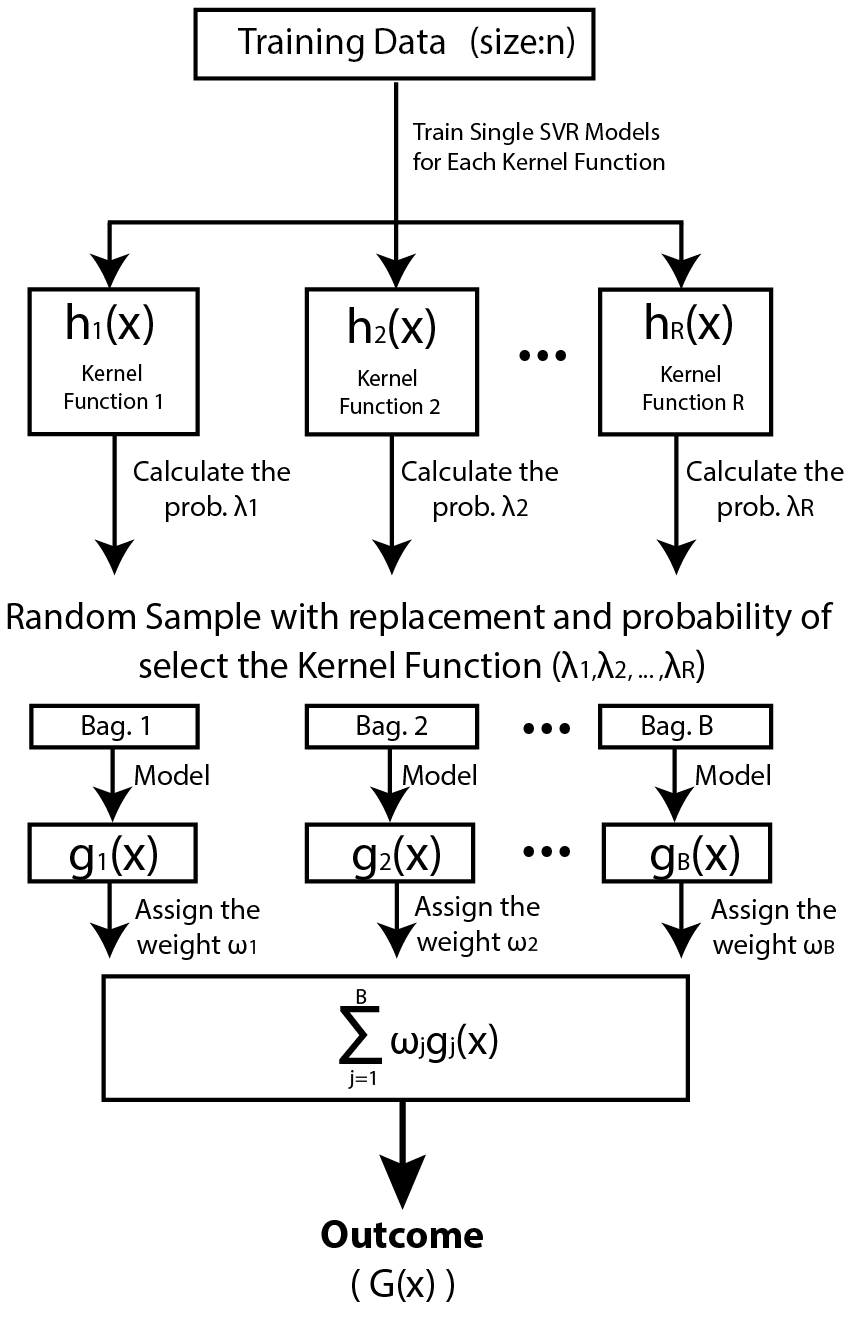}
    \caption{Workflow followed by the Regression Random Machines.}
    \label{fig:sketch}
\end{figure}

The entire Regression Random Machines is schematically presented in Figure \ref{fig:sketch}, where it is designed all the steps used in all cases presented in this article. %The parameter which determines the number of bootstraps samples is chosen by the user, as well as the hyperparameters from the kernel function.

\section{Artificial Data Application}

\begin{figure}[H]
    \centering
    \includegraphics[width=0.3\textwidth]{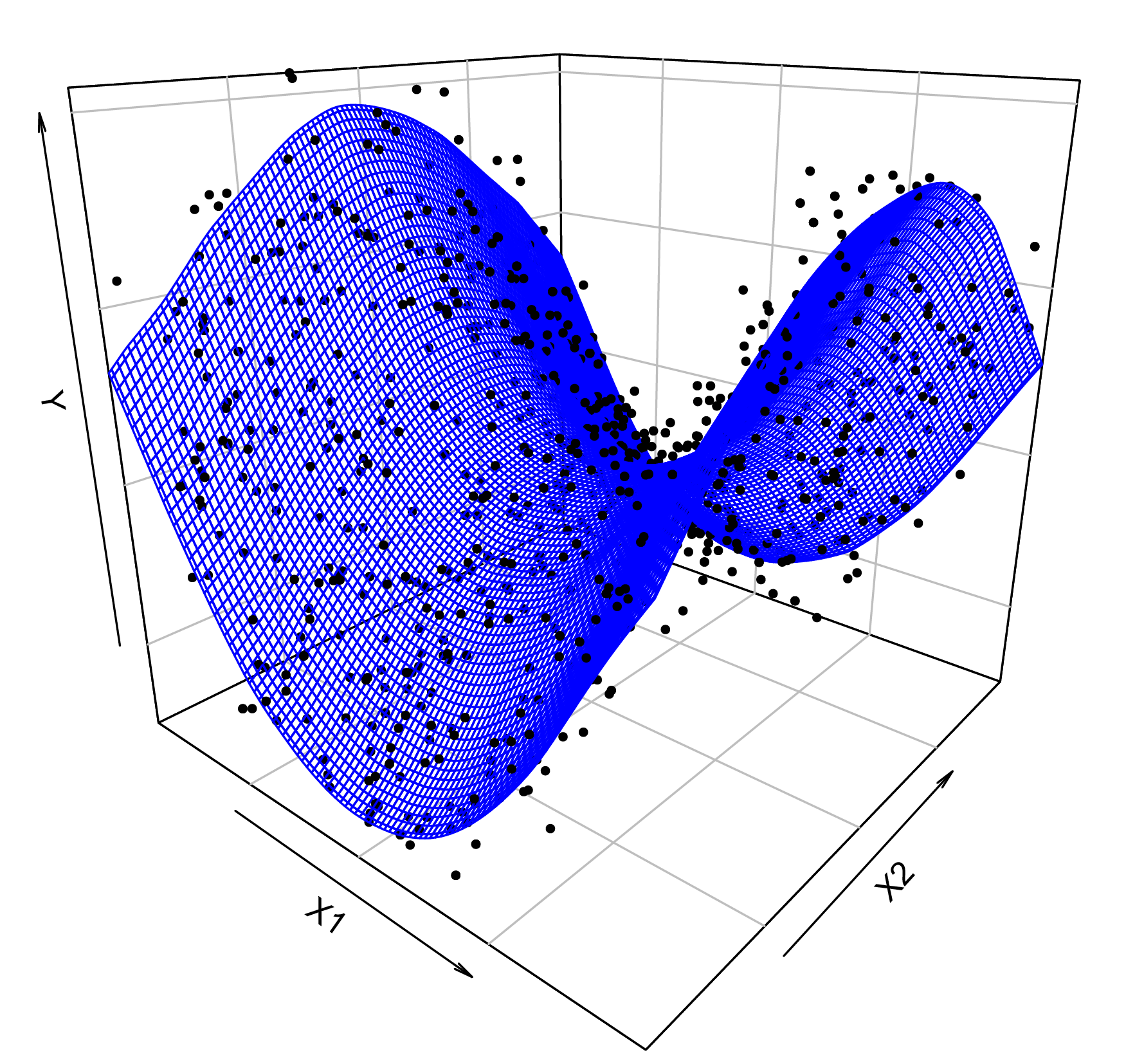}
    \caption{Regression hyperplane of Regression Random Machine algorithm.}
    \label{fig:surface_rrm}
\end{figure}

Different scenarios were used to study the Regression Random Machines on simulated data. The objective was to evaluate the performance and behavior of the model when we have a controlled experiment. Eight different data sets generation scenarios were tested. The \textbf{Models 1-5} are toy examples and can be found in \citep{scornet2016random}, the  \textbf{Model 6} in \citep{van2007super}, the \textbf{Model 7} in \citep{meier2009high} and the \textbf{Model 8} is presented in \citep{roy2012robustness}. The simulations from \textbf{1-7} has the vector of independent predictions $\boldsymbol{X}=(X_{1},\dots,X_{p})$ and $\boldsymbol{X}$ follow a uniform distribution $[0,1]^{p}$. In the \textbf{Model 8} each predictor $X_{i}$ follow a standard normal distribution. Also, we define the transformation $\Tilde{X}_{i}=2(X-0.5),  \hspace{0.2cm}i=1,\dots,p$. For each case the sample size changed among the values of $n =\{30,100,1000\}$.  All the scenarios are described below:

\begin{itemize}
    \item \textbf{Model 1}: p=2, $Y=\Tilde{X^{2}_{1}}+e^{{-\Tilde{X^{2}_{2}}}} + \mathcal{N}(0,0.25)$
    \item \textbf{Model 2}: p=8, $Y=\Tilde{X_{1}}\Tilde{X_{2}}+\Tilde{X^{2}_{3}}-\Tilde{X_{4}}\Tilde{X_{7}}+\Tilde{X_{5}}\Tilde{X_{8}}-\Tilde{X^{2}_{6}}+\mathcal{N}(0,0.5)$
    \item \textbf{Model 3}: p=4, $Y= -sin(\Tilde{X_{1}})+\Tilde{X^{2}_{2}}+\Tilde{X_{3}}-e^{{-X^{2}_{4}}}+\mathcal{N}(0,0.5)$
    \item \textbf{Model 4}: p=4, $Y= \Tilde{X_{1}}+(2\Tilde{X_{2}}-1)^{2}+2sin(2\pi\Tilde{X_{3}})/(2-sin(2\pi\Tilde{X_{3}}))+sin(2\pi\Tilde{X_{4}})+2cos(2\pi\Tilde{X_{4}})+3sin^{2}(2\pi\Tilde{X_{4}})+4cos^{2}(2\pi\Tilde{X_{4}})+\mathcal{N}(0,0.5)$
     \item \textbf{Model 5}: p=8,  $Y=\mathds{1}_{\Tilde{X_{1}} > 0}+ \Tilde{X^{3}_{2}}+\mathds{1}_{\Tilde{X_{3}}+\Tilde{X_{4}}-\Tilde{X_{6}}-\Tilde{X_{5}}  > 1+\Tilde{X_{7}}}+e^{{-\Tilde{X^{2}_{8}}}}+\mathcal{N}(0,0.5)$
    \item \textbf{Model 6}: p=6, $Y=\Tilde{X^{2}_{1}}+\Tilde{X}^{2}_{2}\Tilde{X_{3}}e^{-|\Tilde{X_{4}}|}+\Tilde{X_{6}}-\Tilde{X_{5}}+\mathcal{N}(0,0.5)$
    \item \textbf{Model 7}: p=4, $Y=\Tilde{X_{1}}+3\Tilde{X}^{2}_{2}-2e^{-\Tilde{X_{3}}}+\Tilde{X_{4}}$
    \item \textbf{Model 8}: p=6, $Y=X_{1}+0.707 X^{2}_{2} + 2\mathds{1}_{X_{3}>0}+0.873 \log (X_{1})|X_{3}|+0.894 X_{2} X_{4}+2\mathds{1}_{X_{5}>0}+0.464e^{X_{6}}+\mathcal{N}(0,1)$
    
\end{itemize}
An illustration about how the regression hyperplane created by the RRM can be seen in Figure \ref{fig:surface_rrm} which the \textbf{Model 1} of data generation is used as example.
The repeated holdout with 30 repetitions was used as validation technique with a split ratio of training-test of $70\%-30\%$. The Table \ref{tab:rmse_sim_table} summarizes the result, and from it is possible to realize that the Regression Random Machine outperformed the others methods in the majority simulations setups that were presented. The bold RMSE mean values indicates that this result was the lower among all methods. The default parameters of models were $\gamma=1$, $B=100$, $C=1$ , $\beta=2$, $d=2$ and $\epsilon=0.1$.

% Please add the following required packages to your document preamble:
% \usepackage{multirow}
% Please add the following required packages to your document preamble:
% \usepackage{multirow}
% Table generated by Excel2LaTeX from sheet 'Plan1'
% Table generated by Excel2LaTeX from sheet 'Plan1'
% Table generated by Excel2LaTeX from sheet 'Plan1'

% Table generated by Excel2LaTeX from sheet 'Plan1'

% Table generated by Excel2LaTeX from sheet 'Plan1'
% Table generated by Excel2LaTeX from sheet 'Sheet1'
% Table generated by Excel2LaTeX from sheet 'Sheet1'
\begin{table}[H]
  \centering
  \caption{Summary of the simulation's results for the different databases. Each value corresponds to the mean RMSE method on the test data . In bold the lower value of the generalization error.}
  \begin{adjustbox}{max width=\textwidth}
    \begin{tabular}{ccccccccccc}
    \toprule
    \textbf{Model} & \textbf{n} & \textbf{SVR.Lin} & \textbf{SVR.Pol} & \textbf{SVR.Gau} & \textbf{SVR.Lap} & \textbf{BSVR.Lin} & \textbf{BSVR.Pol} & \textbf{BSVR.Gau} & \textbf{BSVR.Lap} & \textbf{RRM} \\
    \midrule
    \multirow{3}[2]{*}{1} & \multicolumn{1}{r}{30} & 0.4443 & 0.3328 & 0.2189 & 0.2604 & 0.4305 & 0.2942 & 0.2454 & 0.2882 & \textbf{0.2134} \\
          & \multicolumn{1}{r}{100} & 0.4507 & 0.1403 & 0.1482 & 0.1631 & 0.4452 & 0.1370 & 0.1522 & 0.1770 & \textbf{0.1226} \\
          & \multicolumn{1}{r}{1000} & 0.3877 & 0.1110 & 0.1091 & 0.1132 & 0.3876 & 0.1108 & 0.1088 & 0.1120 & \textbf{0.1069} \\
    \midrule
    \multirow{3}[2]{*}{2} & \multicolumn{1}{r}{30} & 1.2274 & 1.4203 & 0.9653 & 0.9435 & 1.2221 & \textbf{0.9146} & 0.9643 & 0.9511 & 0.9191 \\
          & \multicolumn{1}{r}{100} & 1.0425 & 0.7983 & 0.9333 & 0.9016 & 1.0293 & 0.7401 & 0.9379 & 0.9122 & \textbf{0.7390} \\
          & \multicolumn{1}{r}{1000} & 0.8900 & 0.4998 & 0.7996 & 0.6643 & 0.8888 & 0.4983 & 0.8176 & 0.6937 & \textbf{0.4980} \\
    \midrule
    \multirow{3}[2]{*}{3} & \multicolumn{1}{r}{30} & 0.9202 & 2.1038 & 1.0702 & 1.0368 & 0.9181 & 1.5041 & 1.1493 & 1.1121 & \textbf{0.8761} \\
          & \multicolumn{1}{r}{100} & 0.6176 & 1.4123 & 0.8254 & 0.7077 & 0.6116 & 1.3375 & 0.8906 & 0.7738 & \textbf{0.5959} \\
          & \multicolumn{1}{r}{1000} & 0.6086 & 1.2475 & 0.5675 & 0.5373 & 0.6082 & 1.2220 & 0.5612 & 0.5338 & \textbf{0.5334} \\
    \midrule
    \multirow{3}[2]{*}{4} & \multicolumn{1}{r}{30} & 2.2237 & 4.4928 & 2.7105 & 2.6977 & 2.3196 & 3.8646 & 2.8099 & 2.7963 & \textbf{2.1556} \\
          & \multicolumn{1}{r}{100} & 2.2462 & 2.9495 & 2.3342 & 2.1906 & 2.2313 & 2.8389 & 2.3770 & 2.2504 & \textbf{2.1394} \\
          & \multicolumn{1}{r}{1000} & 2.1302 & 3.0750 & \textbf{1.0068} & 1.1613 & 2.1295 & 3.0458 & 1.0734 & 1.2927 & 1.0970 \\
    \midrule
    \multirow{3}[2]{*}{5} & \multicolumn{1}{r}{30} & 0.9664 & 1.8465 & 1.1168 & 1.0756 & \textbf{0.8867} & 1.3307 & 1.1088 & 1.0846 & 0.9025 \\
          & \multicolumn{1}{r}{100} & 0.7740 & 1.9348 & 1.0089 & 0.9074 & \textbf{0.7700} & 1.4642 & 1.0115 & 0.9355 & 0.7757 \\
          & \multicolumn{1}{r}{1000} & 0.7003 & 0.9621 & 0.8835 & 0.6925 & 0.6998 & 0.9561 & 0.8978 & 0.7115 & \textbf{0.6876} \\
    \midrule
    \multirow{3}[2]{*}{6} & \multicolumn{1}{r}{30} & \textbf{0.6806} & 2.7753 & 0.9458 & 0.8629 & 0.7353 & 1.4283 & 0.9517 & 0.8968 & 0.7554 \\
          & \multicolumn{1}{r}{100} & 0.6380 & 1.2054 & 0.9416 & 0.7962 & 0.6395 & 1.1986 & 0.9687 & 0.8440 & \textbf{0.6353} \\
          & \multicolumn{1}{r}{1000} & 0.5792 & 1.0204 & 0.6521 & 0.5491 & 0.5792 & 1.0015 & 0.6709 & 0.5570 & \textbf{0.5453} \\
    \midrule
    \multirow{3}[2]{*}{7} & \multicolumn{1}{r}{30} & \textbf{0.6806} & 2.7753 & 0.9458 & 0.8629 & 0.7353 & 1.4283 & 0.9517 & 0.8968 & 0.7554 \\
          & \multicolumn{1}{r}{100} & 0.6380 & 1.2054 & 0.9416 & 0.7962 & 0.6395 & 1.1986 & 0.9687 & 0.8440 & \textbf{0.6353} \\
          & \multicolumn{1}{r}{1000} & 0.5792 & 1.0204 & 0.6521 & 0.5491 & 0.5792 & 1.0015 & 0.6709 & 0.5570 & \textbf{0.5453} \\
    \midrule
    \multirow{3}[2]{*}{8} & \multicolumn{1}{r}{30} & 2.2623 & 4.2416 & 2.2105 & 2.0928 & 2.0908 & 2.4499 & 2.2323 & 2.1550 & \textbf{2.0283} \\
          & \multicolumn{1}{r}{100} & 1.8324 & 2.5624 & 2.0880 & 1.9245 & 1.8286 & 2.3039 & 2.1121 & 1.9658 & \textbf{1.7994} \\
          & \multicolumn{1}{r}{1000} & 1.9270 & 2.1882 & 1.8161 & \textbf{1.4392} & 1.9252 & 2.1468 & 1.8884 & 1.5254 & 1.5107 \\
    \midrule
    \textbf{Total} & -     & 25.4471 & 42.3213 & 25.8916 & 23.9261 & 25.3307 & 34.4164 & 26.5210 & 24.8669 & \textbf{21.5729} \\
    \bottomrule
    \end{tabular}%
    \end{adjustbox}
  \label{tab:rmse_sim_table}%
\end{table}%

\newpage

\section{Real Data Application}

The methodology was applied on 26 real-world datasets from the UCI Repository \cite{Dua:2019} to evaluate its performance. The datasets present a wide variety in the number of observations, dimensionality, and type of data. In addition, all of them represent a regression task. Table \ref{tab:datasets} summarizes all datasets considered. The continuous features were scaled to zero mean and unit variance, in the exception of the discrete features which were went through a one-hot-encoding process. The validation technique used was the repeated holdout with 30 repetitions and a split ratio of training-test of $70\%-30\%$.

% Table generated by Excel2LaTeX from sheet 'Sheet2'
\begin{table}[H]
\caption{Description of the twenty six regression datasets.}
\begin{adjustbox}{max width=\textwidth}
 \begin{tabular}{clcc|clcc}
    \toprule
    \textbf{ID } & \textbf{Data Set} & \multicolumn{1}{l}{\textbf{\# Instances}} & \multicolumn{1}{l|}{\textbf{\# Features}} & \textbf{ID } & \textbf{Data Set} & \multicolumn{1}{l}{\textbf{\# Instances}} & \multicolumn{1}{l}{\textbf{\# Features}} \\
    \midrule
    1     & \textit{abalone} & 4177  & 7     & 14    & \textit{machines} & 208   & 7 \\
    2     & \textit{airbnb} & 10498 & 17    & 15    & \textit{mpg} & 398   & 6 \\
    3     & \textit{airfoil} & 1502  & 5     & 16    & \textit{ozone} & 330   & 8 \\
    4     & \textit{boston housing} & 505   & 13    & 17    & \textit{parkinson} & 1040  & 26 \\
    5     & \textit{cars} & 50    & 1     & 18    & \textit{petrol} & 31    & 4 \\
    6     & \textit{cement} & 12    & 4     & 19    & \textit{pyrim} & 74    & 27 \\
    7     & \textit{concrete} & 1030  & 8     & 20    & \textit{servo} & 167   & 19 \\
    8     & \textit{cpus} & 208   & 6     & 21    & \textit{slump} & 102   & 7 \\
    9     & \textit{friedman\#1} & 500   & 10    & 22    & \textit{space ga} & 3107  & 6 \\
    10    & \textit{friedman\#2} & 500   & 4     & 23    & \textit{stormer} & 22    & 2 \\
    11    & \textit{friedman\#3} & 500   & 4     & 24    & \textit{taiwan} & 414   & 6 \\
    12    & \textit{geysers} & 298   & 2     & 25    & \textit{triazine} & 185   & 60 \\
    13    & \textit{hills} & 34    & 2     & 26    & \textit{yatch} & 308   & 6 \\
    \bottomrule
    \end{tabular}%
  \label{tab:datasets}%
\end{adjustbox}
\end{table}

\begin{figure}[H]
    \centering
    \includegraphics[width=\textwidth]{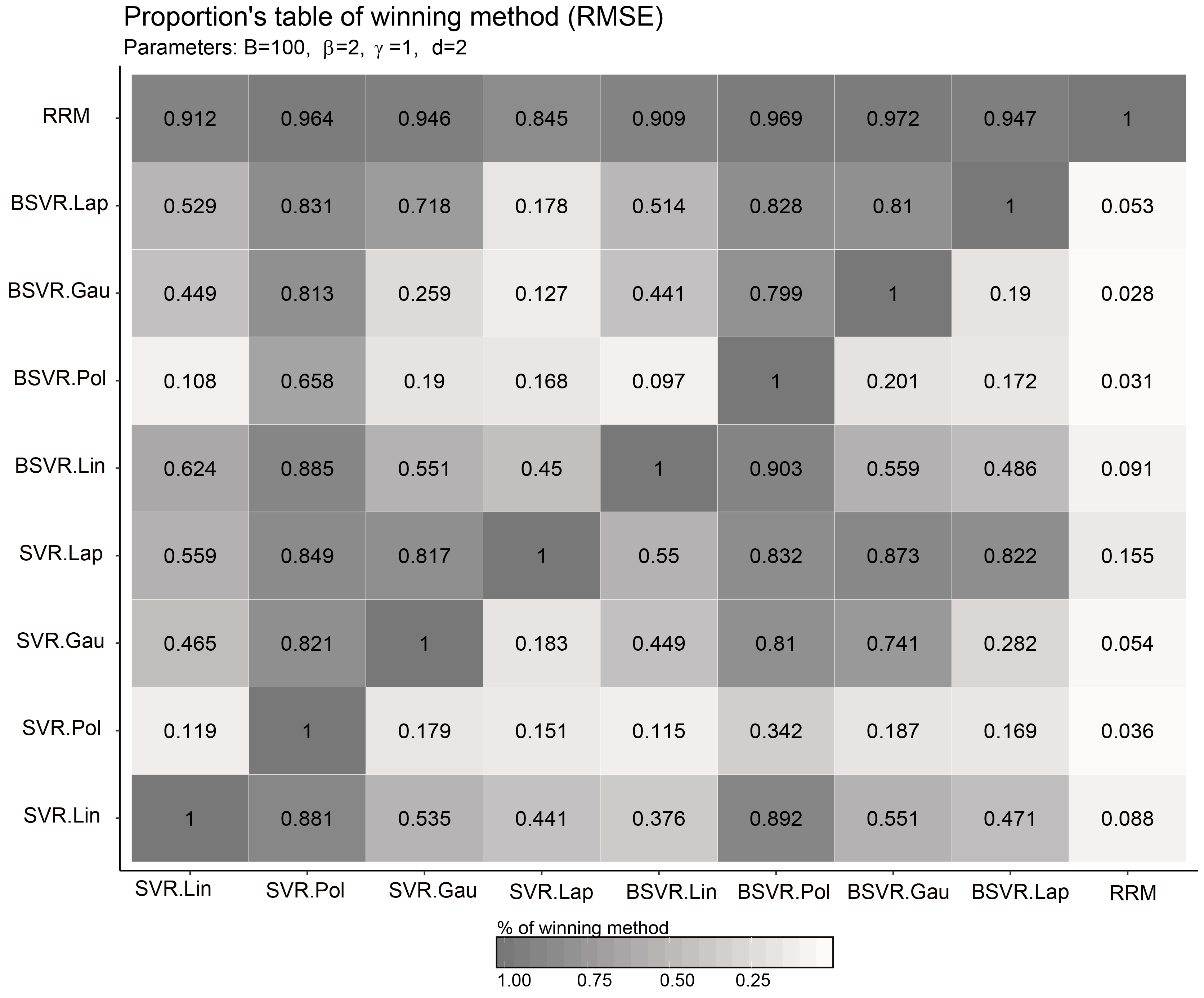}
    \caption{Proportion of the number of times which a method obtained lower RMSE than the others. The proportion summarizes the applications over all 26 datasets and 30 holdout values. It is clear the superiority of the Random Machines when it is compared with the other models. }
    \label{fig:prop_table}
\end{figure}

The Regression Random Machines was compared with the bagged SVR approach using each one of single kernel functions showed in  Table \ref{tab:kernel_rrm}, and with the standard SVR procedure with the same kernel functions. The chosen parameters were: the parameter $\epsilon=1$, the cost parameter $C=1$, the number of bootstrap samples $B=100$, the degree of polynomial kernel $d=2$, and the hyperparameter $\gamma$ from the Table \ref{tab:kernel_rrm}, $\gamma=1$. The result is resumed in the Figure \ref{fig:prop_table} considering the Root Mean Squared Error.

As demonstrated in Figure \ref{fig:prop_table}, the RRM shows lower generalization error than the other bagged support vectors using unique kernel functions. Comparing the RRM with the traditional bagged SVR, it outperforms almost 90.9\% of times the Kernel Linear Bagging, 96.9\% for the Kernel Polynomial Bagging, 97.2\% for the Gaussian Bagging, and 94.7\% for the Laplacian Kernel Bagging. This results shows off that the random weighted choice of the kernels functions reduced,  mostly, the error from the predicted values. The difference is also present when the Regression Random Machines is put on comparison with the singular SVR, where the RM is more accurate $91.2\%$ of times considering the Kernel Linear, $96.4\%$ for the Kernel Polynomial, $94.6\%$ for the Gaussian Bagging, and $84.5\%$ for the Laplacian Kernel.

\begin{figure}[H]
    \centering
    \includegraphics[width=0.7\textwidth]{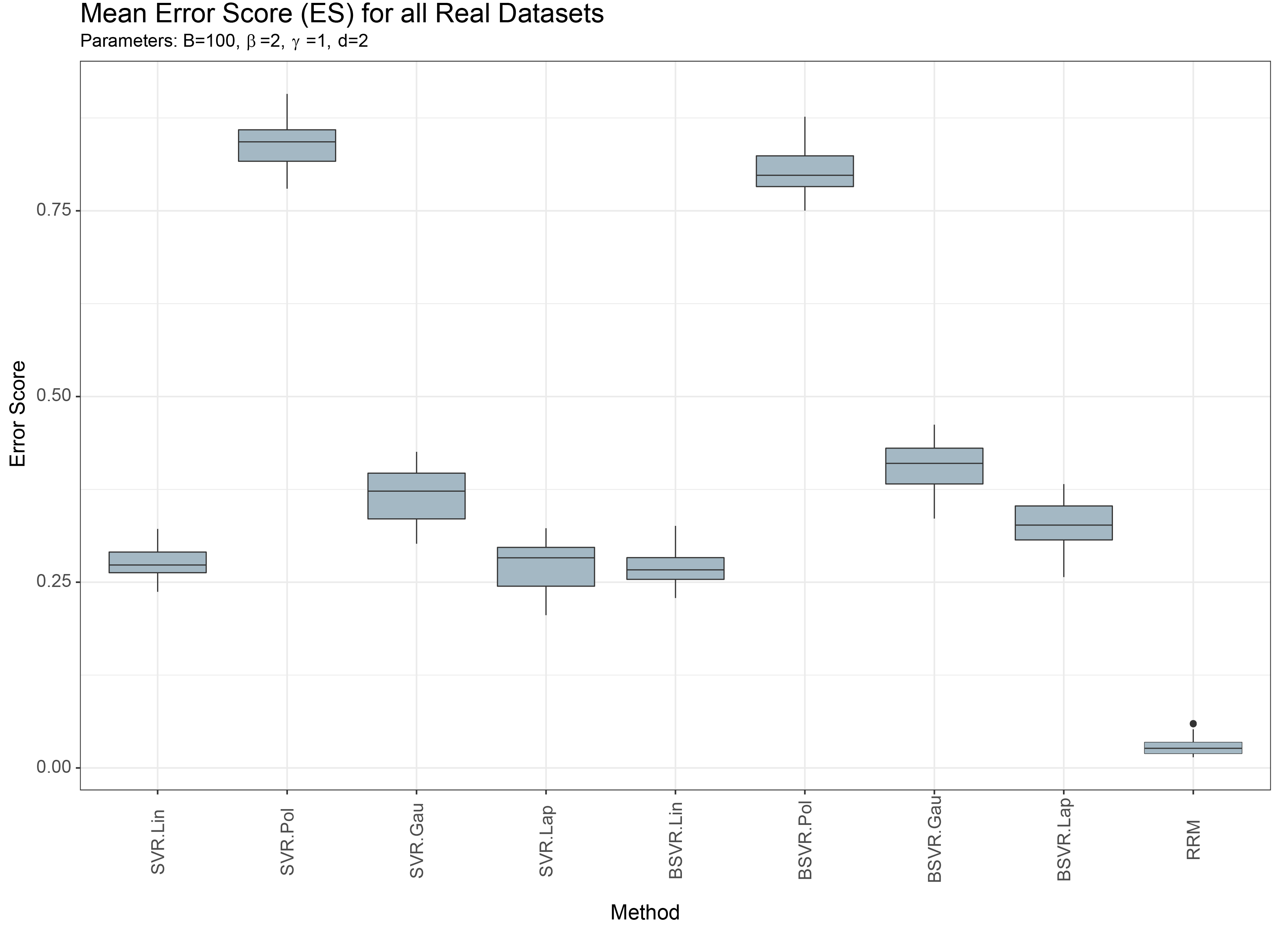}
    \caption{Boxplot of the mean values for each technique used. The result emphasize the general good performance of the Regression Random Machines when compared with the other ones. }
    \label{fig:rmse_score}
\end{figure}

Another way to evaluate the method's performance is through the Error Score (ES) vector. This index formulation it is given by
\begin{equation}
    ES_{i}=\frac{\varepsilon_{i} - \varepsilon_{min}}{\varepsilon_{max}-\varepsilon_{min}}, 
\end{equation}

where the $\boldsymbol{\varepsilon}$ it just the Root Mean Squared Error vector over test observations for all methods, and where $\varepsilon_{i}$ is the individual value for that technique $\forall i=1,\dots,R$. For instance, suppose three algorithms: Regression Random Machines, SVR.Lin Regression and BSVR.Lin, then, after calculating the RMSE for each of them over a test set, the vector $\boldsymbol{\varepsilon}=\{0.1,0.5,0.3\}$ is obtained, with the coordinates for each kernel function respectively. Thus, the $\boldsymbol{ES}$ vector is given by $\boldsymbol{ES}=\{0,1,0.5\}$ which means that $ES_{1}=0$ was the Error Score for RRM, $ES_{2}=1$ for SVR.Lin and $ES_{2}=0.5$ for BSVR.Lin. The Figure \ref{fig:rmse_score} shows boxplots for the mean values for the Error Score over all the 30 holdout repetitions for all 26 benchmarking data sets.

Analyzing the results it is clear to see that the RRM have, generally, a good performance when compared with the traditional methods. This approach also deals with the problem of the choice of best kernel function, since is not necessary to perform a grid-search among all the different functions and define which one has lower test error. For this reason, the RRM algorithm can be considered efficient, as it can reduce the prediction error and the computational cost.

\begin{figure}[H]
    \centering
    \includegraphics[width=0.97\textwidth]{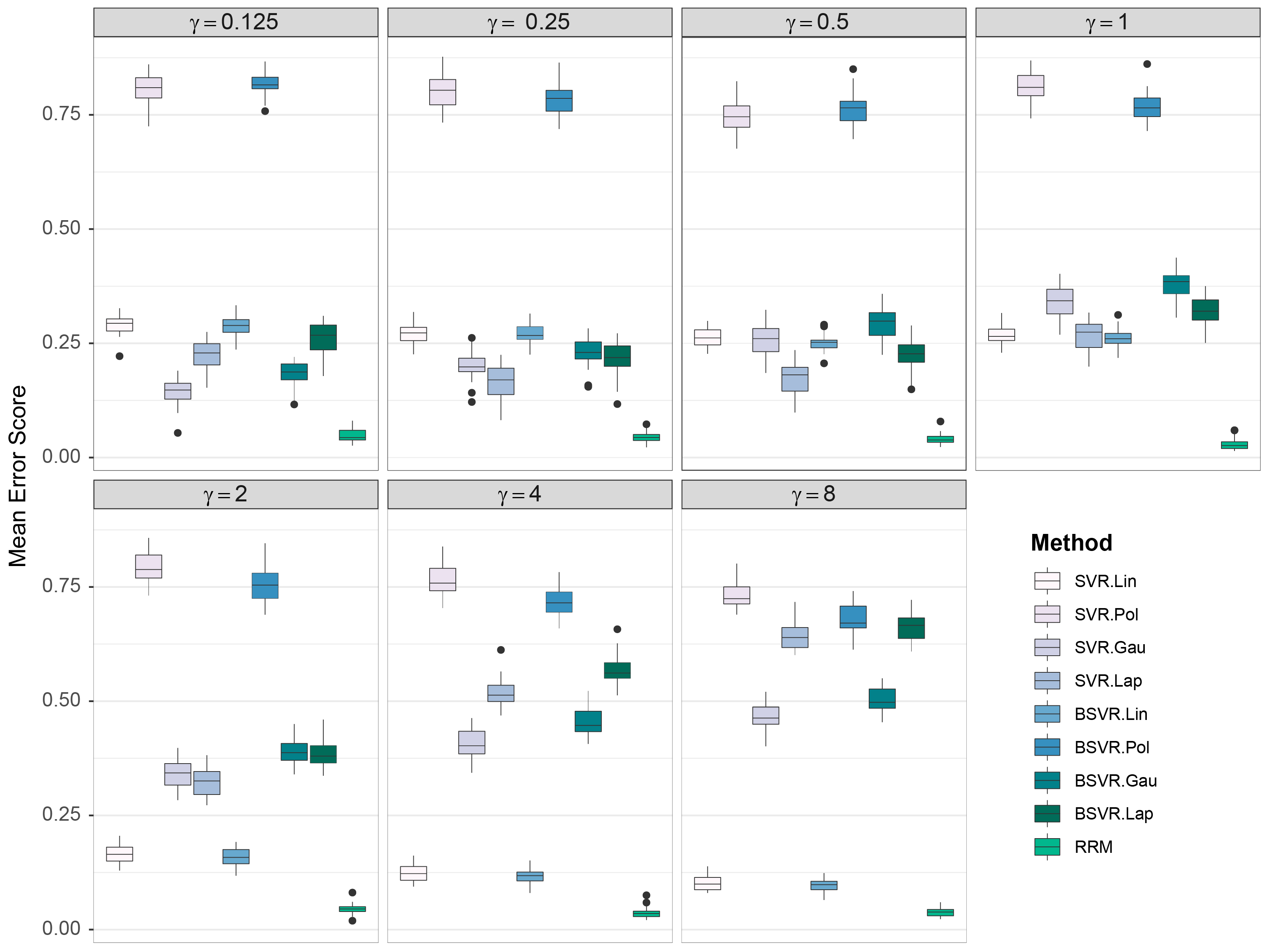}
    \caption{Summary of the mean values for the Error Score applied over 25 real datasets with the different kernel functions and gamma's values. The result reinforces the good performance of RRM despite the modification in hyperparameters settings.}
    \label{fig:rmse_gamma}
\end{figure}

As the hyperparameter tuning is an important procedure in of the support vector machine regression algorithm \cite{duan2003evaluation}, the value of $\gamma$ was changed in order to study how its variation affects the behavior of RRM. The setting of the parameters was the set of values $\gamma=\{2^{-3},2^{-2},2^{-1},2^{0},2^{1},2^{2},2^{3}\}$ over the same 25 data sets (removed the \textit{Airbnb}). The result is shown in Figure \ref{fig:rmse_gamma}, where it is possible to see that the RRM surpassed the other bagging and single models. As said before the selection of these hyparameters, as the kernel function, has a direct impact on the model performance, and the results fortify the supposition that RRM gives a good and consistent result for a wide range $\gamma$ values.

\section{Strength and Correlation evaluation}

In this section we explain the reason why the Regression Random Machines approach is an ensemble approach that can reduce the generalization error. The random selection of kernel functions works to diversify different functions that belong to a Reproducing Kernel Hilbert Space (RHKS). The goal of this procedure is to diminish the correlation between regression models that constitute the RRM and increase strength of them since these both components result in greater results to bagged classifiers \citep{breiman2001random}. 

 The correlation concept can be defined as a measure of how much models are similar, while the strength of a model relies on how well it correctly predicts an observation. The estimation of a correlation measure can have different approaches. Considering classification models, for instance, a method to estimate the correlation between models is to calculate the area from decision boundaries that overlaps among them \cite{turney1995bias}. Other estimation method, still considering classification context, is used by \citep{timkamho1998random}, who defines the similarity, through the agreement measure, as the number of observations that are equally labeled with the same class by different models.
 
 In the regression approach the correlation/similarity estimation between models can be calculated as the mean of the upper triangle from the correlation matrix given in Equation \ref{eq:cor_matrix}.
 
 \begin{equation}
\Sigma_{corr} = 
\begin{pmatrix}
    \rho_{\hat{y}_{1},\hat{y}_{1}} & \rho_{\hat{y}_{1},\hat{y}_{2}} & \ldots & \rho_{\hat{y}_{1},\hat{y}_{B-1}} & \rho_{\hat{y}_{1},\hat{y}_{B}} \\
     & \rho_{\hat{y}_{2},\hat{y}_{2}} & \ldots & \rho_{\hat{y}_{2},\hat{y}_{B-1}} & \rho_{\hat{y}_{2},\hat{y}_{B}} \\
     & & \ddots & \vdots & \vdots \\
     & & & \rho_{\hat{y}_{B-1},\hat{y}_{B-1}} & \rho_{\hat{y}_{B-1},\hat{y}_{B}} \\
     &  &  &  & \rho_{\hat{y}_{B},\hat{y}_{B}}
\end{pmatrix} _{B\times B}
\label{eq:cor_matrix}
\end{equation}
 
 The values of $\rho_{\hat{y}_{i},\hat{y}_{j}}$ are calculated by
 
 \begin{equation}
     \rho_{i,j}={\frac {\sum _{k=1}^{T}(\hat{y}_{i_{k}}-\overline{\hat {y_{i}}  })(\hat{y}_{j_{k}}-\overline{\hat {y_{j}}  })}{{\sqrt {\sum_{k=1}^{T}(\hat{y}_{i_{k}}-\overline{\hat {y_{i}}  })}}{\sqrt {\sum _{k=1}^{T}(\hat{y}_{j_{k}}-\overline{\hat {y_{j}}  })}}}}.
 \end{equation}

for all $i\neq j=1,\dots,B$, and $\hat{y_{i}}$ it is the vector of predictions from observations that belongs to the test set. The strength in this article will be estimated using the Error Score presented in the previous Section 3.6, since it capture well the prediction performance and has the same range of the correlation measure. As the ES is directly proportional to the Root Mean Square Error, it also can be considered a strength measure. Smalls values of RMSE produced by a regression model implies in a stronger model. 

In order to assess the correlation and strength of the RRM in comparison with the traditional bagged version of SVR, the algorithm was applied over all models of simulated data presented at the Section 3.5. We look for the model which has the lowest RMSE and Error Score (i.e: greater strength) \textbf{and} the lowest correlation measure. A model with small agreement can benefit more from the bagging procedure \cite{breiman2001random}. However, just small values of correlation are not enough, since this lower value can represent a weak model, i.e, a model which is not capable to predict new observations well. The result is summarized in Table \ref{tab:acc_agr}. Both, RMSE and Agreement were calculated using a 30 Repeated Holdout validation set with split ratio of 70-30\% training-test. The parameters of the methods were: B=100, $\gamma=1$, C=1, $\beta=2$ and $\epsilon=0.1$.

The strength of the models is affected by the agreement and vice-versa, so optimize both measures at same time is a difficult effort. The relation between them can be analyzed in the Table \ref{tab:acc_agr}. Observing simultaneously the RMSE and the Agreement measure from traditional bagging approaches exists a traded-off between them. Considering a data set as example Table \ref{tab:acc_agr}, if the RMSE is the lower among them, its agreement is the highest. This trade-off is minimized in RRM case, which, despite presents the lower RMSE in most of the cases, this is not reflected in the highest agreement measure among all methods. Therefore,  Regression Random Machines has the low correlation and great strength, desirable features to produce a good bagging approach.

\begin{figure}[H]
    \centering
    \includegraphics[width=\textwidth]{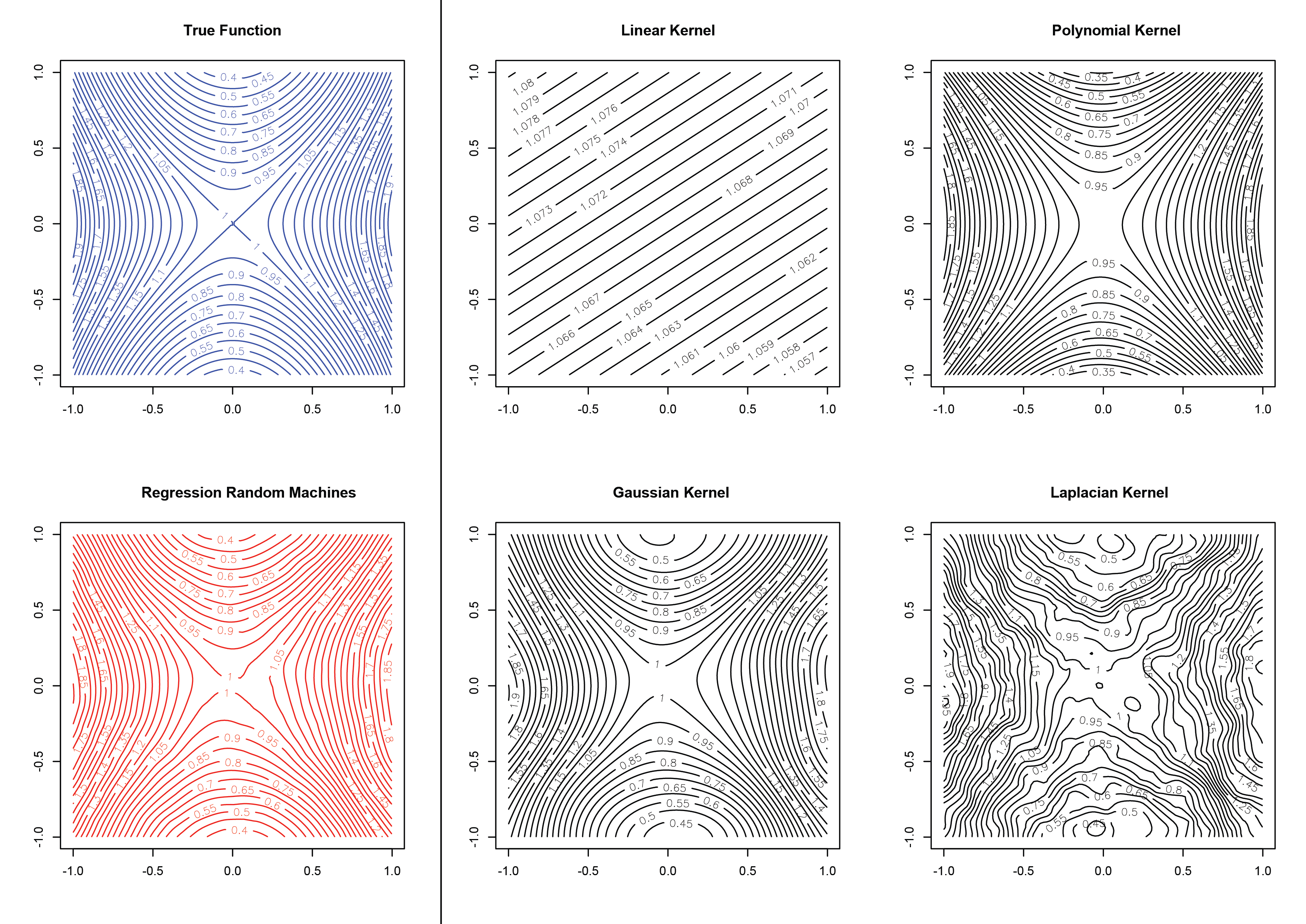}
    \caption{Level Curves for different single Kernel Function models used in the Regression Random Machines. It is important to analyse the diversity between the different kernel functions, and how the RRM surface is the closest to the True Function. }
    \label{fig:kernel_grid}
\end{figure}

% Table generated by Excel2LaTeX from sheet 'Sheet1'
\begin{table}[htbp]
  \centering
  \caption{Summary of Strength (RMSE) and Agreement.}
    \begin{adjustbox}{max width=0.8\textwidth}
    \begin{tabular}{cccccccccccc}
    \toprule
    \textbf{Model} & \textbf{n} & \multicolumn{2}{c}{\textbf{BSVR.Lin}} & \multicolumn{2}{c}{\textbf{BSVR.Pol}} & \multicolumn{2}{c}{\textbf{BSVR.Gau}} & \multicolumn{2}{c}{\textbf{BSVR.Lap}} & \multicolumn{2}{c}{\textbf{RRM}} \\
    \midrule
          &       & RMSE  & AGR   & RMSE  & AGR   & RMSE  & AGR   & RMSE  & AGR   & RMSE  & AGR \\
    \midrule
    \multirow{3}[2]{*}{1} & \multicolumn{1}{r}{30} & 0.1189 & 0.6427 & 0.0812 & 0.2417 & 0.0773 & 0.6146 & 0.0777 & 0.7342 & 0.0725 & 0.5236 \\
          & \multicolumn{1}{r}{100} & 0.0538 & 0.9531 & 0.0216 & 0.3337 & 0.0196 & 0.7873 & 0.0246 & 0.9163 & 0.0147 & 0.8806 \\
          & \multicolumn{1}{r}{1000} & 0.0117 & 0.9953 & 0.0049 & 0.5841 & 0.0028 & 0.9187 & 0.0038 & 0.9797 & 0.0030 & 0.8822 \\
    \midrule
    \multirow{3}[2]{*}{2} & \multicolumn{1}{r}{30} & 0.2972 & 0.5300 & 0.2388 & 0.4839 & 0.2209 & 0.8472 & 0.2229 & 0.8329 & 0.1913 & 0.7263 \\
          & \multicolumn{1}{r}{100} & 0.1356 & 0.2617 & 0.1086 & 0.9340 & 0.1361 & 0.9620 & 0.1394 & 0.9648 & 0.1287 & 0.9303 \\
          & \multicolumn{1}{r}{1000} & 0.0272 & 0.1630 & 0.0214 & 0.9926 & 0.0292 & 0.9954 & 0.0284 & 0.9916 & 0.0213 & 0.9276 \\
    \midrule
    \multirow{3}[2]{*}{3} & \multicolumn{1}{r}{30} & 0.1788 & 0.5754 & 0.5976 & 0.4217 & 0.2672 & 0.5810 & 0.2644 & 0.7254 & 0.1983 & 0.4719 \\
          & \multicolumn{1}{r}{100} & 0.0641 & 0.8014 & 0.1715 & 0.1746 & 0.1260 & 0.6121 & 0.1183 & 0.8669 & 0.0749 & 0.6634 \\
          & \multicolumn{1}{r}{1000} & 0.0279 & 0.9821 & 0.0443 & 0.5348 & 0.0253 & 0.8171 & 0.0245 & 0.9642 & 0.0252 & 0.8622 \\
    \midrule
    \multirow{3}[2]{*}{4} & \multicolumn{1}{r}{30} & 0.5890 & 0.7763 & 1.4439 & 0.2917 & 0.6318 & 0.7119 & 0.6273 & 0.7904 & 0.4998 & 0.6229 \\
          & \multicolumn{1}{r}{100} & 0.1671 & 0.9142 & 0.4010 & 0.5741 & 0.3106 & 0.8236 & 0.2577 & 0.9065 & 0.2199 & 0.7995 \\
          & \multicolumn{1}{r}{1000} & 0.0608 & 0.9927 & 0.1172 & 0.8581 & 0.0551 & 0.9700 & 0.0408 & 0.9833 & 0.0515 & 0.9223 \\
    \midrule
    \multirow{3}[2]{*}{5} & \multicolumn{1}{r}{30} & 0.2044 & 0.5607 & 0.2242 & 0.2609 & 0.1664 & 0.5638 & 0.1647 & 0.7267 & 0.1630 & 0.4967 \\
          & \multicolumn{1}{r}{100} & 0.0811 & 0.8381 & 0.1487 & 0.3304 & 0.0956 & 0.7468 & 0.0958 & 0.8854 & 0.0743 & 0.7168 \\
          & \multicolumn{1}{r}{1000} & 0.0215 & 0.9854 & 0.0333 & 0.8492 & 0.0267 & 0.9249 & 0.0242 & 0.9788 & 0.0225 & 0.9141 \\
    \midrule
    \multirow{3}[2]{*}{6} & \multicolumn{1}{r}{30} & 0.1813 & 0.6427 & 0.3004 & 0.2417 & 0.1397 & 0.6146 & 0.1411 & 0.7342 & 0.1373 & 0.5236 \\
          & \multicolumn{1}{r}{100} & 0.0641 & 0.9531 & 0.1553 & 0.3337 & 0.1219 & 0.7873 & 0.1181 & 0.9163 & 0.0655 & 0.8806 \\
          & \multicolumn{1}{r}{1000} & 0.0206 & 0.9953 & 0.0429 & 0.5841 & 0.0289 & 0.9187 & 0.0239 & 0.9797 & 0.0226 & 0.8822 \\
    \midrule
    \multirow{3}[2]{*}{7} & \multicolumn{1}{r}{30} & 0.1813 & 0.8227 & 0.3004 & 0.2767 & 0.1397 & 0.6822 & 0.1411 & 0.7825 & 0.1373 & 0.6855 \\
          & \multicolumn{1}{r}{100} & 0.0641 & 0.9774 & 0.1553 & 0.4250 & 0.1219 & 0.8694 & 0.1181 & 0.9516 & 0.0655 & 0.8941 \\
          & \multicolumn{1}{r}{1000} & 0.0206 & 0.9974 & 0.0429 & 0.6676 & 0.0289 & 0.9539 & 0.0239 & 0.9852 & 0.0226 & 0.9113 \\
    \midrule
    \multirow{3}[2]{*}{8} & \multicolumn{1}{r}{30} & 0.3852 & 0.3062 & 0.5018 & 0.3884 & 0.4003 & 0.5006 & 0.4072 & 0.6054 & 0.4057 & 0.3633 \\
          & \multicolumn{1}{r}{100} & 0.3463 & 0.3791 & 0.3931 & 0.3763 & 0.3732 & 0.6027 & 0.3760 & 0.6778 & 0.3612 & 0.4037 \\
          & \multicolumn{1}{r}{1000} & 0.1405 & 0.4648 & 0.1593 & 0.9603 & 0.1609 & 0.7719 & 0.1511 & 0.9143 & 0.1496 & 0.8568 \\
    \midrule
    \textbf{Total} & -     & 3.4430 & 17.5107 & 5.7094 & 12.1191 & 3.7061 & 18.5776 & 3.6149 & 20.7939 & 3.1283 & 17.7414 \\
    \bottomrule
    \end{tabular}%
    \end{adjustbox}
  \label{tab:acc_agr}%
\end{table}%
\newpage
The idea of how the random selection of kernel functions can increase the diversity of Regression Random Machines models, when compared with the traditional SVR bagging algorithm, can not be so clear at the first moment. In order to observe clearly how it works, \textbf{Model 1} will be used, with n=1000, as study case and see graphically the modelling process. Figure \ref{fig:kernel_grid} shows the level curves from the true data generation surface, the predicted hyperplane for RRM, and each single bootstrap model of a single kernel function. It can be seen that for each different kernel function the regression surface is distinct. which promotes diversity and subsequent reduction of correlation between models. Moreover, it is possible to notice that the Regression Random Machines surface, built through the combination of these single models, is the one that is closer to the real data generation hyperplane, reinforcing that it works as the best model in that case.

\newpage

\begin{figure}
    \centering
    \includegraphics[width=\textwidth]{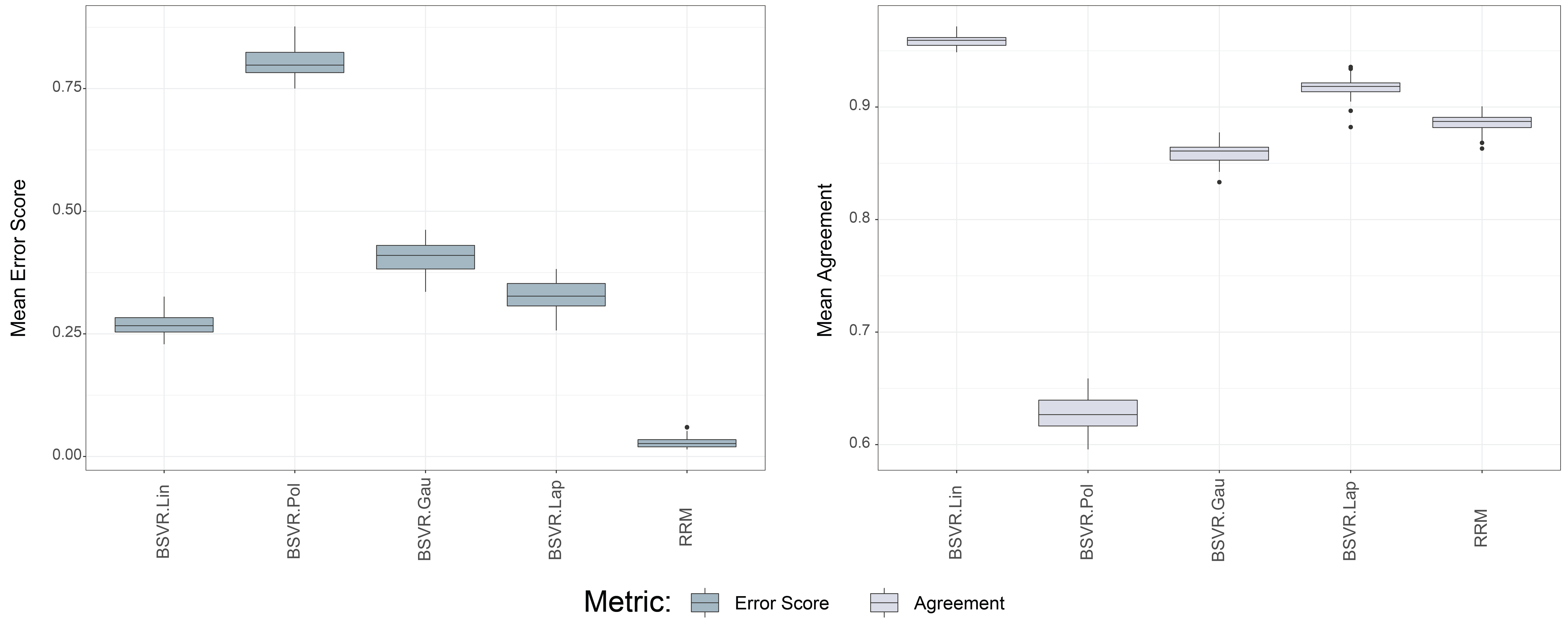}
    \caption{Boxplots of the mean Error Score and Mean Agreement for each method.}
    \label{fig:acc_and_agr}
\end{figure}

The same behavior was observed on the 26 real data sets, where the agreement is also calculated and compared with the strength (Error Score) of each model (Figure \ref{fig:acc_and_agr}). Although the low values of generalization error from BSVR.Lin and BSVR.Lap they present large agreement values. On other hand, despite the low values of agreement from BSVR.Pol and BSVR.Gau they produce great values of Error Score. The unique method that can perform the optimal values for both it is the Regression Random Machines.

\subsection{The Correlation Coefficient $\beta$}

Another way to study the correlation-strength trade-off in the RRM procedure is through the coefficient $\beta$ presented in the Equations \ref{eq:prob_rrm}-\ref{eq:weight_rrm}. As mentioned before, the $\beta$ cofficient can calibrate the diversity of kernel functions used during the bagging procedure. If we consider $\beta=0$ the RRM will hold that all kernel functions will be sampled and weighted equally. On other hand, if we use greater values of $\beta$ the RRM's behavior will be close to traditional SVR' bagging, since just a single kernel function will be chosen.

\begin{figure}[H]
    \centering
    \includegraphics[width=\textwidth]{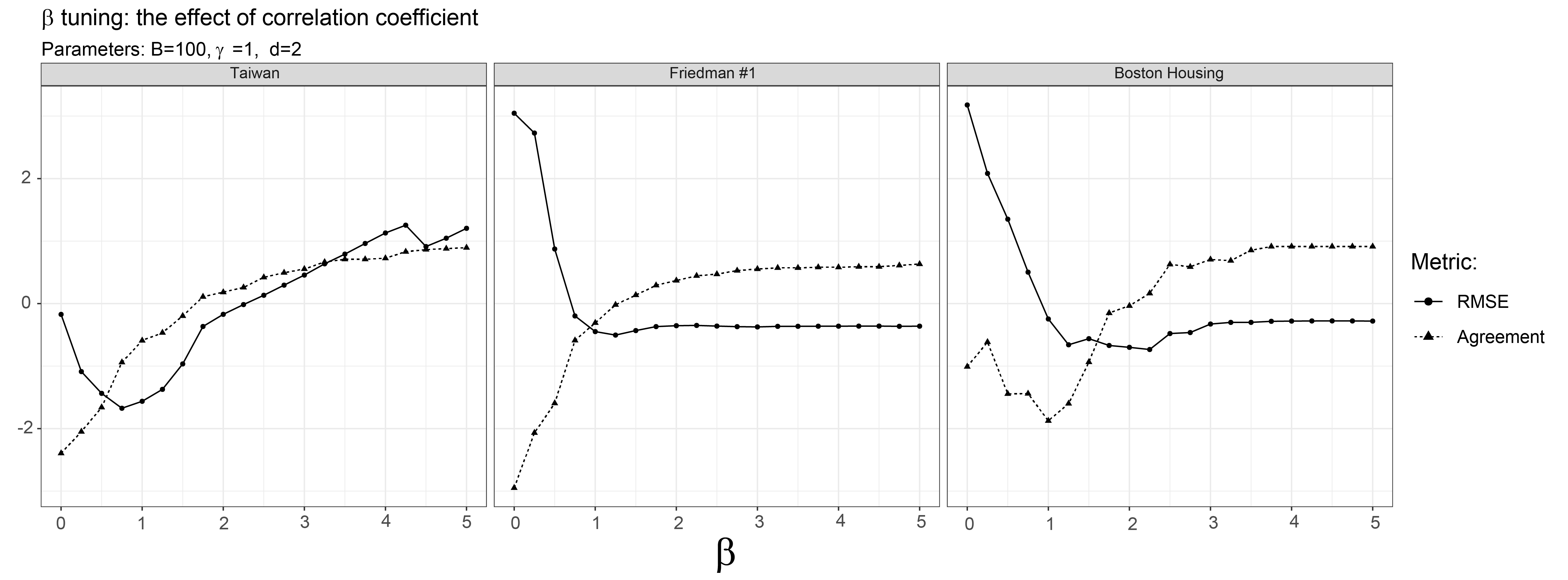}
    \caption{Standardized values of RMSE and Agreement for the different values of $\beta$.}
    \label{fig:beta_tuning}
\end{figure}

To demonstrate this performance, we evaluate the standardized RMSE and Agreement on  benchmarks from Section 3.6 changing the values of $\beta$ in a grid that range from 0 to 5, with the length of 21 intervals.
Both measures were calculated in a holdout validation with split ratio of 70-30\%, and setting the parameters $B=100$, $\gamma=1$ and $d=2$. The results are summarised in three Data Sets: \textit{Taiwan}, \textit{Boston Housing} and \textit{Friedman \#1} presented in Figure \ref{fig:beta_tuning}.

Though different values of RMSE and Agreement, all of them present the same behavior: for small values of $\beta$, i.e: minimum correlation between models, we have an weaker (represented through the large RMSE values) model from RRM algorithm. As the beta value increases the weighting on the kernel functions and bootstrap models predictions are applied increasing the agreement and reducing the RMSE. However, at some inflection point the RMSE starts to increase and the agreement continues to grow. The reason for this result can be explained by \citep{breiman1996bagging}, where it is defined that stable base-models, SVR models in this case, will not benefit from the aggregation procedure, and may even depreciate the model. Large values of $\beta$ starts to penalize the RRM in way that just one, or few different,  kernel functions will be chosen, and as SVR \cite{breiman1996heuristics} is defined as a stable model (i.e: bootstrap replications produces small changes in the model) this may lead to worst results. Therefore, the key of the improvement from Regression Random Machines is to add the instability in SVR, necessary for bagging procedures \cite{breiman1996bagging}, through the random sample of kernel functions and the weights associated to the predictions.

The $\beta$ also can be defined as a hyperparameter of the model and can be tuned in order to achieve the lowest generalization error. From some empirical results, in this article the default choice of $\beta$ was 2, since in the most of cases the best choice have been around this value and there wasn't much improvement from the grid search procedure for this parameter.

\newpage 

\section{Final Comments}

The main contribution of this paper is to propose a novel learning approach to do ensemble using Support Vector Regression models that can enhance the improve the traditional bagging SVR, and give an alternative to deal with the problem choosing the best kernel function that should be used. Through the Regression Random Machines, the combination of different SVR  models by the different kernel functions propose an algorithm that avoids the expensive computational cost of doing a grid or random search between the kernel functions, besides reduce the general prediction error. 

In order to quantify this reduction, suppose an number of $B$ models calculated in a traditional bagging procedure and $R$ as the number of kernels functions that will be evaluated and used in support vector models. In traditional bagging algorithm using SVR as base-models the number of total models that will be calculated to obtain the bests results is given by $B \times K$, while using the Regression Random Machines approach this number reduces to $B + K$. Using an example of $B=100$ and $K=400$, we have that the traditional bagging algorithm would take approximately four times the computational cost than the proposed Random Machines since the ratio of calculated models is 400/104 (i.e: four time faster).

Furthermore, the results from RRM explored the strength and correlation characteristics in the bagging procedure, obtaining simultaneously lower generalization error and agreement, instead of traditional ensemble procedures using SVR as base models which cannot obtain them at the same time.

Despite the success of Regression Random Machines there are some open problems. For instance, although the RRM avoids the kernel function choice, the tuning for some hyperparameters of SVR still necessary for achieve the best model, and considering the number of models in bagging, the tuning can be computationally expensive. Additionally, this methodology can be explored in other contexts, as the computational cost, and can be applied to any practical statistical learning problem. Future theoretical studies may be done with respect to the use of other and more kernel functions in the bagging step, besides other weighting functions approaches.

%% The Appendices part is started with the command \appendix;
%% appendix sections are then done as normal sections
%\appendix

%\label{appendix-sec1}

%% Sample text. Sample text. Sample text. Sample text. Sample text. Sample text. 
%% Sample text. Sample text. Sample text. Sample text. Sample text. Sample text. 
%% Sample text. 

%% References
%%
%% Following citation commands can be used in the body text:
%% Usage of \cite is as follows:
%%   \citep{key}         ==>>  [#]
%%   \cite[chap. 2]{key} ==>> [#, chap. 2]
%%

%% References with bibTeX database:

%% The Appendices part is started with the command \appendix;
%% appendix sections are then done as normal sections
%\appendix

%\label{appendix-sec1}

%% Sample text. Sample text. Sample text. Sample text. Sample text. Sample text. 
%% Sample text. Sample text. Sample text. Sample text. Sample text. Sample text. 
%% Sample text. 

%% References
%%
%% Following citation commands can be used in the body text:
%% Usage of \cite is as follows:
%%   \cite{key}         ==>>  [#]
%%   \cite[chap. 2]{key} ==>> [#, chap. 2]
%%

%% References with bibTeX database:
\section*{Acknowledgements}

I would like to thank CAPES-CNPq, for financing this project and providing the necessary financial resources for the complete this research.

%% References with bibTeX database:
%\section*{References}
\bibliography{mybibfile}

\end{document}